\documentclass[sigconf,table,dvipsnames]{acmart}

\usepackage{booktabs}
\usepackage{array}
\usepackage{multirow}
\usepackage{enumitem}
\usepackage{longtable}
\usepackage{tikz}
\usepackage{amsmath}
\usepackage{float}
\usepackage{placeins}
\usetikzlibrary{shapes.geometric,arrows,positioning,calc}

\renewcommand\footnotetextcopyrightpermission[1]{}
\AtBeginDocument{\providecommand\BibTeX{{\normalfont B\kern-0.5em{\scshape i\kern-0.25em b}\kern-0.8em\TeX}}}
\graphicspath{{./}}
\begin{document}
\title{Integrating Fairness and Explainability in a Multiple Instance Reinforcement Learning System}

\author{Bente Hinkenhuis}
\affiliation{\institution{University of Amsterdam}\city{Amsterdam}\country{The Netherlands}}
\email{bente.hinkenhuis@student.uva.nl}

\author{Seyed Sahand Mohammadi Ziabari}
\authornote{Corresponding author.}
\affiliation{\institution{University of Amsterdam}\city{Amsterdam}\country{The Netherlands}}
\email{s.s.mohammadiziabari@uva.nl}

\author{Ali Mohammed Mansoor Alsahag}
\affiliation{%
  \institution{University of Amsterdam}
  \city{Amsterdam}
  \country{The Netherlands}
}
\email{a.m.m.alsahag@uva.nl}

\begin{abstract}
Predicting student performance from educational interaction data requires models that are both accurate and sufficiently transparent to support meaningful intervention, while demographic information creates an additional risk of unfair predictions. This study investigates a multi-objective framework that combines Reinforcement Learning-based Multiple Instance Learning (RL--MIL), adversarial debiasing, and preference-conditioned hypernetworks for student-at-risk prediction. MIL represents each student as a bag of weakly labeled interactions, while an RL agent selects informative instances for downstream classification. Two hypernetwork variants are evaluated to determine whether a user-defined preference scalar can continuously control the trade-off between predictive performance and Equalized Odds. The underlying RL--MIL baseline achieves strong classification performance, but both hypernetwork extensions exhibit mode collapse: changing the preference weight produces little systematic movement along the intended fairness--performance frontier. The failure is associated with objective dominance, weak gradient propagation through the conditioning mechanism, and interactions between dynamically generated parameters. The results show that fairness objectives can be incorporated into an interpretable RL--MIL pipeline, but preference conditioning alone does not guarantee controllable multi-objective behavior. Robust fair RL--MIL therefore requires explicit mechanisms for gradient balancing, objective separation, and stability analysis.
\end{abstract}
\keywords{Multiple Instance Learning, Reinforcement Learning, Adversarial Debiasing, Student-at-Risk Prediction, Fairness}
\fancyhead{}
\maketitle

\section{Introduction}
\label{sec:introduction}
Multiple Instance Learning (MIL) has developed into a broadly applicable weakly supervised learning paradigm, with applications ranging from biomedical analysis to text and educational data mining \cite{waqas_exploring_2024}. In educational settings, MIL is particularly attractive because student outcomes are often observed at the level of an entire course trajectory while the individual interactions that drive those outcomes remain unlabeled.

One such area is student performance prediction. When a student successfully completes or fails a course, we often have rich information about them. That includes demographic information such as age and gender, but also their grades in all the assessments during the course. In some datasets, there are even recordings of their interactions with an online learning platform. However, we do not know which of these data are most predictive of a student passing or failing. We can only assign a label to the bag of information as a whole. This key characteristic of missing labels for individual data-points but available labels for sets of data is a textbook example for a MIL problem. A MIL classifier can be trained to predict whether a student passes or fails based on the already collected bag of information about them.

As helpful as it seems, prediction of the final outcome alone is not enough. Particularly for deriving helpful interventions, it is crucial to know which information is most indicative of the predicted outcome. If a student was troubled by a certain kind of assignment, this potential intervention would look very different from one that would be designed for a student not interacting with certain kinds of online resources. Thus, we are interested in transparent instance selection.

Instance selection techniques strive to discard repetitive or uninformative instances in a bag to sharpen the focus on the informative ones. This improves the predictions of a MIL classifier down the line. Previous work has shown a wide variety of such techniques. One promising direction is to use a Reinforcement Learning (RL) agent for this task. This approach allows dynamic instance selection and reward functions targeted at multiple objectives. Recent RL--MIL work on student outcomes further shows that context-aware attention policies can preserve predictive performance while making the selection mechanism intrinsically interpretable \cite{braakman_improving_nodate}. This flexibility is central to the present study because the selection policy must be evaluated not only for predictive utility, but also for demographic fairness and the way those objectives reshape which student interactions are retained.

The study therefore evaluates whether fairness-focused adversarial debiasing can be integrated into an interpretable RL--MIL framework for student-at-risk prediction without eroding predictive performance or destabilizing the instance-selection policy. The analysis focuses on two coupled aspects. First, it examines the attainable trade-off between classification performance and Equalized Odds under preference-conditioned multi-objective optimization, including whether the resulting configurations form a meaningful Pareto front. Second, it tests whether stronger debiasing pressure changes the learned selection policy away from static demographic information and toward assessment and interaction evidence, while preserving selection sparsity and consistency. Together, these analyses assess whether a hypernetwork-conditioned RL--MIL architecture can provide controllable fairness--performance behavior rather than a single fixed operating point.

\section{Related Work}
\label{sec:related_work}
Tackling fairness in interpretable RL-based instance selection requires careful attention to the known issues of the key aspects of this problem. An approach that optimizes for information density in their selected instances easily suffers from biases when demographic metadata is available for prediction. Adversarial Debiasing and Causal Regularization offer two differently focused ways to alleviate this shortcoming.

In this section, different approaches to instance selection using RL are reviewed, followed by a summary of several ways of improving interpretability for RL-based methods. Then, challenges in fair RL are discussed. Lastly, key concepts related to causal modeling in RL are examined.
\subsection{Instance Selection using RL}
Performing dynamic adaptation to task-specific rewards, RL-based instance selection has proven to be a valuable tool. Contrary to statistical methods which are computationally expensive because of all the different possible instance combinations and do not allow for online learning, a reinforcement learning algorithm can learn the implications of selecting a combination of instances autonomously. Adapting to new data or different settings (bigger bags for example) also does not require a full retraining of the model, but can be learned by running another set of episodes instead (online learning) \cite{zhou_reinforcement_2022}. Despite its promising results, it also shows several areas that need improvement. One of these is its ability to select the most informative instances. \cite{benjamins_instance_2024} highlights that RL-based instance selection models over-prioritize frequently occurring but less informative instances, thereby failing to generalize across datasets and exhibiting biased learning. For Curious Instance Selection (CIS) and a reward function balancing data reduction and accuracy, \cite{moran_curious_2022} shows that there is no guarantee for the most valuable instances to be selected. This issue arises particularly when prioritizing accuracy gains over generalization.

Stability in the selection process is also challenging to achieve in many settings. Especially during early training, convergence is hindered by random exploration and delayed rewards. Recent extensions of RL--MIL illustrate both its flexibility and its optimization sensitivity. Lakeman et al.~\cite{lakeman_multitask_2026} extend reinforcement-driven instance selection to multi-task speaker-attribute prediction and show that shared representations can improve multiple weakly supervised tasks simultaneously, while the cross-language RLMIL-DAT framework combines RL-based selection with domain-adversarial training to suppress language-specific cues \cite{shu_crosslanguage_2026}. These studies are methodologically relevant because they demonstrate that additional objectives and adversarial signals can be incorporated into RL--MIL, but they also motivate careful evaluation of transfer, stability, and objective interference. In a medical context, \cite{lin_sg-murcl_2026} demonstrates that combining RL-based instance selection with a Graph Neural Network (GNN) and an attention smoothing operator makes the instance selection policy collapse entirely, always proposing trivial solutions. \cite{fan_feature_2022} states that their dual-agent RL framework required manual tuning for stability, since it suffered from action space explosion.

Instance selection using RL techniques offer several benefits when their drawbacks are mitigated by handling biases and designing training paradigms carefully. One of these benefits is a natural way to assess the interpretability of the instance selection policy.
\subsection{Interpretability of RL Methods}
In their review of explainable RL, \cite{puiutta_explainable_2020} define interpretability as the ability to generate, extract and present explanations that are understandable to humans, allowing them to predict the model's behavior. For any predictive modeling technique, there is a fundamental trade-off between model performance and explanation quality. Simpler models often sacrifice model performance for straightforward explanations, while larger so-called black-box models are harder to reason about. Methods for interpretation extraction are broadly classified into two categories. Intrinsic explainability describes models that include a mechanism for interpretation themselves. This can mean employing a simple prediction process or containing a part that generates explanations at important positions during the prediction process. On the other hand, post-hoc methods provide insight into the prediction characteristics by comparing input-output combinations that are different in a meaningful way \cite{holzinger_explainable_2018}.

For predicting student performance, both approaches have been explored. \cite{boujmiraz_predicting_2026} note that as model complexity grows, the focus on interpretability wanes. Very few of these studies investigate interpretability using post-hoc methods like SHAP or LIME. Also, \cite{glanois_survey_2024} stress that there is a need for standardized interpretability metrics to be able to compare different techniques. A directly related study by Braakman et al.~\cite{braakman_improving_nodate} evaluates attention-based RL--MIL policies for student outcome prediction on OULAD. Their results show that gated and multi-head attention policies can match the predictive performance of an epsilon-greedy RL--MIL baseline while reducing explanation latency and exposing the selected interactions intrinsically. This establishes a close methodological baseline for the present work, which retains interpretable instance selection but adds fairness as an explicit optimization objective.

Interpretable models for student performance prediction help educators to derive suitable interventions based on a student's need. Particularly in an educational context, however, it is crucial that these models' predictions are as unbiased as possible.
\subsection{Fairness in RL}
Fair Machine Learning techniques aim to reduce bias in model outcomes. \cite{rabonato_systematic_2025} classify these into pre-processing, in-processing and post-processing based approaches. Pre-processing means reducing bias in the data on which a model is trained by for example balancing sample amounts for the combinations of protected features and class labels. In-processing, in turn, encompasses all frameworks aiming to combat bias in the model itself, for example by introducing an adversarial debiasing agent punishing biased classifications. Lastly, post-processing based interventions alter the decisions made based on the model's output, by for example varying the decision threshold depending on the value of a protected feature \cite{rabonato_systematic_2025}.

In the domain of student performance prediction, several approaches have been explored. As an example of a pre-processing technique, \cite{sha_lessons_2023} reports improved fairness metrics for models trained on a student performance dataset that was previously balanced using Class-Balancing Techniques (CBT). Latent-space mitigation has likewise been studied specifically for at-risk student prediction, illustrating that demographic disparity can persist even when overall predictive performance is strong \cite{al-zawqari_latent_2024}. The widely used approach of in-processing bias mitigation using adversarial debiasing has also been applied to student admissions \cite{priyadarshini_fair_2024}. Beyond education, task-adaptive adversarial debiasing has shown that the strength of a debiasing signal can be varied across instances to preserve task utility where stereotypical evidence is weak \cite{zhu_taskadaptive_2025}; this is conceptually aligned with the present attempt to control the fairness--performance trade-off through a preference-conditioned objective.

Reducing bias of classification models for student performance prediction is possible in a variety of ways. However, aggregate performance and a single disparity score can conceal heterogeneous failure patterns. Recent fairness-metric work shows that global model quality may mask substantial label- or domain-specific disparities and argues for diagnosis-oriented fairness evaluation alongside standard performance metrics \cite{naranbat_fairness_2025}. This supports the present use of Equalized Odds together with predictive performance and policy-level diagnostics rather than treating fairness as a single isolated objective. Still, spurious correlations, proxy variables, and unobserved confounding can limit purely associational debiasing methods. Recent work comparing adversarial and causal disentanglement similarly finds that suppressing demographic leakage can introduce a substantial utility trade-off \cite{olijslager_causally_2026}. These limitations motivate the causal regularization directions discussed later in this paper.

\section{Methodology}
\label{sec:methodology}
This work extended the work of \cite{braakman_improving_nodate}. Therefore, the corresponding architecture of their efforts will serve as a baseline to all comparisons made.

Student performance prediction can be formulated as a MIL problem:

\begin{align}
\mathcal{D}=\{(B_i,y_i)\}_{i=1}^N    
\end{align}.

Every $B_i=\{x_{i1},\dots,x_{iM}\}$ denotes the super-bag for student $i$ with each $x_{ij}$ corresponding to a single instance. $y_i \in \{0, 1\}$ stands for the label of that student, indicating whether they successfully passed the course (indicated by a $1$) or are considered at-risk because of failure or drop out (indicated by a $0$).

Out of these super-bags, subsets $b_i \subset B_i$ with much smaller magnitude $|b_i| \ll |B_i|$ were constructed during the instance selection phase to increase model efficiency, performance and interpretability. For each bag $b_i$, the final classifier output a single prediction $\hat{y}_i$.

Since we examine model fairness, we further defined an index set $I_P=\{i|1\leq i\leq M\}$ showing which instances of a super-bag are protected features where potential bias should be mitigated. The protected features of a student $i$ would thus be:

\begin{align}
P_i=\{x_{ij}|x_{ij} \in B_i \land j \in I_P\}
\end{align}

\subsection{Data}
The Open University Learning Analytic Dataset (OULAD) as first published by \cite{kuzilek_open_2017} contains information about enrolled student's participation in seven modules and their demographic information. Besides their demographic information and registration details, it provides insight into assessment results and their participation in a virtual learning environment (VLE), operationalized as the number of clicks per day on a particular learning material.

\subsection{Preprocessing}
As it has been shown to be beneficial for model performance by \cite{braakman_improving_nodate}, we constructed sets consisting of the (1) static features of each student, their (2) assessment results and (3) their interactions with the VLE aggregated by activity type using summation. In order to not eliminate temporal information during this aggregation, the three columns first click day, last click day and number of click days were added to the aggregated data.

\textit{Numerical normalization}: In order to keep the influence of every attribute on the model's learning equal, all numerical features were scaled to a range of $[0, 1]$ using Min-Max scaling.

\textit{Categorical feature encoding}: After transforming categorical features into integer indices, we constructed their dense embeddings via a learnable layer.

\textit{Padding and Masking}: The final MIL model requires fixed-size tensors. Thus, we padded every bag to the maximum number of instances (39). During training, we used a binary mask to distinguish between real instances and padding.

\subsection{Exploratory Data Analysis}
To develop hypotheses for what to expect from trying to debias the instance selection and/or classification, we calculated a simple correlation matrix. This way, we saw which variables in the data were associated to each other and were able to reason about which data would probably be downweighted because of their relatedness to protected features and which protected features' exclusion from bag selections would probably lead to the most significant performance decline.

\subsection{Baseline model}
Since \cite{braakman_improving_nodate} found that the framework introducing an RL agent based on the one described by \cite{salkhordeh_ziabari_reinforced_2024} exhibited the most sparse instance selection policy, we based our framework on that structure. An RL agent selected instances out of the super-bags and fed the selected instances to a MIL classifier, which in turn output the label prediction. This prediction was used for backpropagation based on cross-entropy in the baseline framework, updating the classifier's and the RL agent's weights using the optimistic Adam algorithm.

The MIL classifier is composed of a pretrained autoencoder, that transforms instances into their low-dimensional vector representation. These vectors are then pooled into one vector for the whole bag based on one of four pooling schemes. These include mean pooling, max pooling, attention-based \cite{ilse_attention-based_2018} and Rep the Set \cite{skianis_rep_2020} pooling. As a last step, a multi-layer perceptron predicts the label for the given vector representation.

\subsection{Proposed multi-objective framework}
In our extended framework, we introduced a second RL agent that predicts four protected demographic features of each student based on the outputs of the last hidden layer of the multi-layer perceptron in the MIL classifier. For each of the student's gender, age, hometown and educational background, one MLP was used for prediction. The predicted values were then transformed into their cross-entropy using the values of the corresponding instances and the mean cross-entropy was used as the fairness signal.

\textit{Pareto set learning}: Based on this signal and the performance signal from the predicted label (MIL classifier), the Pareto-front optimization scheme described by \cite{liu_pareto_2025} was used to explore the trade-off between prediction accuracy and bias. A simple MLP output a set of parameters $\theta_2$ for a given preference configuration drawn from a uniform distribution that was mixed with the parameters $\theta_1$ of the instance selection agent and/or the task model MLP trained on the classification reward signal according to this formula:
\begin{align}
\theta = (1-\alpha)\theta_1+\alpha\theta_2,
\end{align}
where the weighting parameter $\alpha$ was adjusted during hyperparameter selection. More concretely, during an episode we drew $k \in \mathbb{N}$ preference weight vectors $w \in [0,1]$ uniformly that reflect the targeted balance between classification performance $w=0$ and bias detectability $w=1$. This scalar value was transformed into a Fourier Embedding to account for the complexity hidden behind this one-dimensional value. Proper control of the parameters for instance selection (and classification) needs to account for complex nonlinear patterns between different preference configurations that are not reflected by a scalar value. This idea originally stems from computer vision research \cite{tancik_fourier_2020}, but fits our use case well. For every one of these configurations, the hypernetwork yielded a set of parameters to mix into the instance selection agent. The reward and fairness signals of running the parameterized model and the MIL classification on $n \in \mathbb{N}$ super-bags were stored in a shared replay buffer. Based on the runs stored in this replay buffer we stochastically backpropagated on all networks, which concludes an epoch. Parameter optimization was carried out using the optimistic Adam algorithm. We ran two different hypernetwork variants.

\textit{Task-agnostic hypernetwork}: For this variant, the MIL classifier and the debiasing agent were only adjusted based on the classification signal and the fairness signal respectively, as we strived to debias the framework at the instance selection level, not at the classification level. This lets us gain insight into which data are crucial to exclude to protect sensitive features easily. However, it is easily possible that the model's performance collapses if the debiaser is still able to guess protected features, since the task model is not aware of any debiasing in this setup.

\textit{Task-aware hypernetwork}: To account for possible mode collapse, we also trained a hypernetwork that controls both the policy network and the MLP of the task model. This should protect against mode collapse, as protected features can be excluded at the selection level and obfuscated at the classification level.

\begin{figure*}[ht]
\includegraphics[width=\textwidth]{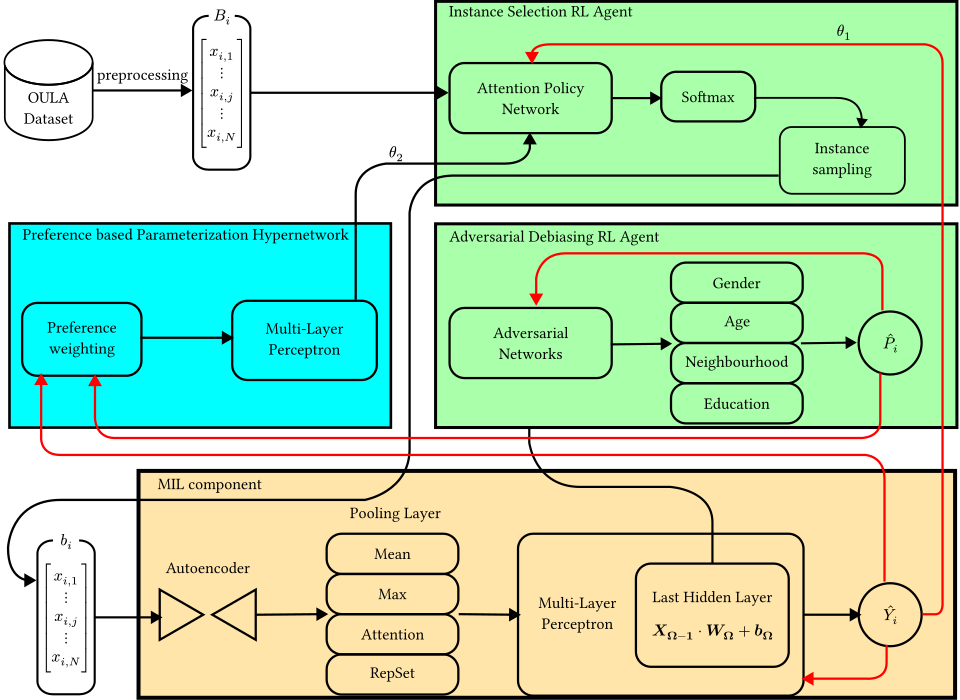}
\Description{A flowchart showing how the two reinforcement learning agents for instance selection and adversarial debiasing, the MIL classifier and a preference based parameterization hypernetwork are connected and how data flows through the architecture.}
\caption{Overview of proposed novel architecture. Black arrows indicate data/parameter flow. Red arrows indicate backpropagation.}
\label{fig:flow}
\end{figure*}

A schematic depiction of the task-agnostic framework can be seen in Figure \ref{fig:flow}. For the task-aware hypernetwork, the only difference is that the MLP of the MIL component has its own set of $theta_1$ parameters that get mixed with $theta_2$ parameters from the hypernetwork as well.

\subsection{Evaluation}
We evaluated our framework in the three explored areas of classification performance, interpretability and fairness. The Pareto-front generated during training was compared to the baseline framework on a test-set using the following metrics:
\begin{itemize}
    \item F1-score for classification performance measured per prediction
    \item Maximum difference between sensitivity and specificity compared across each of the four protected features to reflect equality of odds 
    \item Gini coefficient for explanation sparsity of the instance selection policy
    \item Spearman's rank correlation on two randomly sampled half-splits of the test set and variance across seeds for consistency of the instance selection policy
\end{itemize}
All of these were measured for each Pareto-optimal weight setting accumulated over super-bags (students).

\subsection{Experimental setup}

We conducted experiments to assess model performance and bias towards protected features for 3 different architectures across 4 pooling methods. To assess reproducibility and calculate consistency metrics, we ran every combination of pooling method and architecture on 5 different seeds (0-4). Our results consist of the mean and standard deviation across the seeds. The data was split into a training (80\%), validation (10\%) and test (10\%) set using a weighted random sampler to have an approximately equal number of positive and negative samples in each batch. A consistent bag size of 20 was used to ensure that performance differences could only be attributed to the quality and biasedness of chosen samples, not to the size of a bag. As a first step, we performed hyperparameter selection using the bayesian optimization algorithms provided by Weights \& Biases, since Bayesian algorithms are efficient for finding optimal hyperparameters \cite{wu_hyperparameter_2019}.

\section{Results}
\label{sec:results}
We present the results of the exploratory data analysis first. Then, we compare the performance of each of the three frameworks and its reproducibility for each of the four pooling methods mean pooling, max pooling attention pooling and rep-the-set pooling in this order. Lastly, we present the differences of the interpretability metrics between models.

\subsection{Exploratory Data Analysis}

\begin{figure}[ht]
\includegraphics[width=\columnwidth]{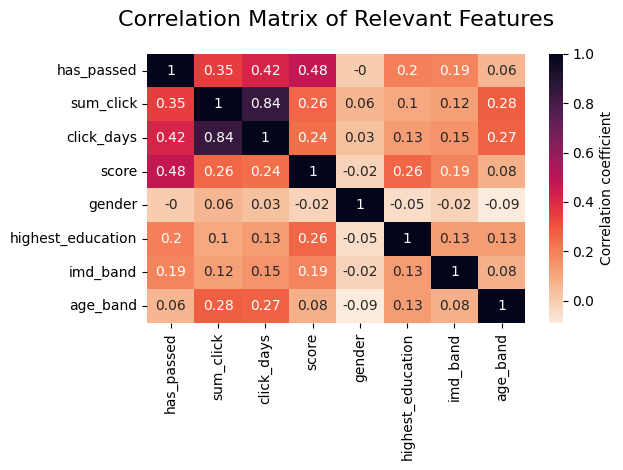}
\Description{TODO}
\caption{Simple correlation matrix calculated using the technically incorrect assumption of interval scaling for each categorical variable to simplify correlation calculation. The relationship between assessment score or click statistics and has\_passed is comparatively the strongest (excluding the obviously high correlation between sum\_click and click\_days). Correlations involving protected features are low, but tend to be higher for scores and click statistics than for has\_passed, indicating that the causal model proposed in the methodology section could have some merit.}
\label{fig:cormat}
\end{figure}

As can be confirmed by examining Figure \ref{fig:cormat}, the protected variables of highest previous education and IMD score weakly correlate to the final outcome of the course and the assessment scores. Thus, we expected that model performance would drop when these features are excluded from selected bags and were therefore less likely to be dropped than the age and gender features. As the age of a person seems to relate to their VLE usage behaviours, we expected that VLE instances would be less likely to be selected for preferences reflecting high bias-awareness. The last protected feature of a person's age seemed to not be related to any other variable, so we expected this to be the first feature to be dropped from selected bags.

\subsection{Model performance analysis}

\begin{table*}[ht]
    \centering
    \begin{tabular}{l|l|r|r|r}
        \textbf{Framework}         & \textbf{Pooling} & \textbf{Preference} & \textbf{F1 score (performance) ($\pm$ SD)} & \textbf{Equalized Odds (bias) ($\pm$ SD)} \\
        \hline
        Baseline                   & Attention        &                     & 0.9245 ($\pm$ 0.0018)                      & 0.1681 ($\pm$ 0.0044)                     \\
        Baseline                   & Max              &                     & 0.9187 ($\pm$ 0.0012)                      & 0.0998 ($\pm$ 0.0288)                     \\
        Baseline                   & Mean             &                     & 0.9228 ($\pm$ 0.0017)                      & 0.1055 ($\pm$ 0.0160)                     \\
        Baseline                   & Rep-The-Set      &                     & 0.9280 ($\pm$ 0.0010)                      & 0.1426 ($\pm$ 0.0237)                     \\
        Task-agnostic Hypernetwork & Attention        & 0.0                 & 0.9261 ($\pm$ 0.0011)                      & 0.1344 ($\pm$ 0.0007)                     \\
        Task-agnostic Hypernetwork & Attention        & 1.0                 & 0.9244 ($\pm$ 0.0020)                      & 0.1356 ($\pm$ 0.0176)                     \\
        Task-agnostic Hypernetwork & Max              & 0.0                 & 0.9196 ($\pm$ 0.0011)                      & 0.1458 ($\pm$ 0.0142)                     \\
        Task-agnostic Hypernetwork & Max              & 1.0                 & 0.9206 ($\pm$ 0.0024)                      & 0.1454 ($\pm$ 0.0006)                     \\
        Task-agnostic Hypernetwork & Mean             & 0.0                 & 0.9207 ($\pm$ 0.0012)                      & 0.1316 ($\pm$ 0.0156)                     \\
        Task-agnostic Hypernetwork & Mean             & 1.0                 & 0.9222 ($\pm$ 0.0014)                      & 0.1630 ($\pm$ 0.0151)                     \\
        Task-agnostic Hypernetwork & Rep-The-Set      & 0.0                 & 0.9270 ($\pm$ 0.0031)                      & 0.1603 ($\pm$ 0.0193)                     \\
        Task-agnostic Hypernetwork & Rep-The-Set      & 1.0                 & 0.9250 ($\pm$ 0.0020)                      & 0.1721 ($\pm$ 0.0065)                     \\
        Task-aware Hypernetwork    & Attention        & 0.0                 & 0.9343 ($\pm$ 0.0002)                      & 0.1705 ($\pm$ 0.0002)                     \\
        Task-aware Hypernetwork    & Attention        & 1.0                 & 0.9351 ($\pm$ 0.0009)                      & 0.1748 ($\pm$ 0.0066)                     \\
        Task-aware Hypernetwork    & Max              & 0.0                 & 0.9098 ($\pm$ 0.0006)                      & 0.1701 ($\pm$ 0.0012)                     \\
        Task-aware Hypernetwork    & Max              & 1.0                 & 0.9056 ($\pm$ 0.0006)                      & 0.1705 ($\pm$ 0.0017)                     \\
        Task-aware Hypernetwork    & Mean             & 0.0                 & 0.9259 ($\pm$ 0.0022)                      & 0.1701 ($\pm$ 0.0008)                     \\
        Task-aware Hypernetwork    & Mean             & 1.0                 & 0.9256 ($\pm$ 0.0015)                      & 0.1727 ($\pm$ 0.0040)                     \\
        Task-aware Hypernetwork    & Rep-The-Set      & 0.0                 & 0.9261 ($\pm$ 0.0020)                      & 0.1788 ($\pm$ 0.0010)                     \\
        Task-aware Hypernetwork    & Rep-The-Set      & 1.0                 & 0.9234 ($\pm$ 0.0014)                      & 0.1658 ($\pm$ 0.0004)                     \\
    \end{tabular}
    \caption{Shortened list of performance metrics for the best models. Metrics for preference values between 0 and 1 are similar to those shown, which can be confirmed by looking at Table \ref{tab:full_performance} in Appendix \ref{sec:apx:B}.}
    \label{tab:short_performance}
\end{table*}

All models exhibit similar classification performance across random seeds and preference values (see Table \ref{tab:all_seeds_performance} in Appendix \ref{sec:apx:B}). Pairwise t-tests for each combination of the three frameworks (at preference value 0) shown in Table \ref{tab:perf_ttests} of Appendix \ref{sec:apx:B} confirm that there is no clear pattern across pooling methods. Only for attention pooling, the task-aware hypernetwork framework shows a significantly better performance than both other models. Similarly, the Equalized Odds Deviation shows no significantly best framework for any pooling method (see Table \ref{tab:bias_ttests} of Appendix \ref{sec:apx:B}.

\begin{table*}[ht]
    \centering
    \begin{tabular}{l|l|l|r|l}
        \textbf{Pooling} & \textbf{Framework}         & \textbf{Tested Hypothesis}  & \textbf{T value} & \textbf{p value} \\
        \hline
        Mean             & Task-agnostic Hypernetwork & Preference 0 > Preference 1 & -0.1782          & 0.8630           \\
        Mean             & Task-aware Hypernetwork    & Preference 0 > Preference 1 & 0.1859           & 0.8697           \\
        Max              & Task-agnostic Hypernetwork & Preference 0 > Preference 1 & -0.3635          & 0.7257           \\
        Max              & Task-aware Hypernetwork    & Preference 0 > Preference 1 & 6.8235           & 0.0208 (*)       \\
        Attention        & Task-agnostic Hypernetwork & Preference 0 > Preference 1 & 0.8337           & 0.4286           \\
        Attention        & Task-aware Hypernetwork    & Preference 0 > Preference 1 & -1.2058          & 0.3512           \\
        Rep-The-Set      & Task-agnostic Hypernetwork & Preference 0 > Preference 1 & 0.9615           & 0.3644           \\
        Rep-The-Set      & Task-aware Hypernetwork    & Preference 0 > Preference 1 & 1.0213           & 0.3370           \\
    \end{tabular}
    \caption{T test of classification F1-score between preference 0 and preference 1 for both hypernetwork architectures and all pooling methods.}
    \label{tab:pareto_performance}
\end{table*}

Comparing between the two extreme preference values of 0 and 1 for each hypernetwork architecture and pooling method shows that the models do not adapt to the preference scalar. For the classification F1-score displayed in Table \ref{tab:pareto_performance}, only one comparison (max pooling, task-aware hypernetwork) is statistically significant and for the Equalized Odds Deviation displayed in Table \ref{tab:pareto_bias} of Appendix \ref{sec:apx:B}, no comparison is statistically significant. The overall low standard deviations show that these results are robust across seeds.

\subsection{Model interpretability analysis}

\begin{table*}[ht]
    \centering
    \small
    \setlength{\tabcolsep}{3.5pt}
    \begin{tabular}{r|r|r}
        \textbf{Baseline}                       & \textbf{Task-agnostic Hypernetwork, preference 0} & \textbf{Task-agnostic Hypernetwork, preference 1} \\
        \hline
        Age Band (98.9774 \%)                   & VLE Clicks: homepage (97.1877 \%)                 & VLE Clicks: homepage (97.4434 \%)                 \\
        Code Presentation (91.8919 \%)          & Assessment: TMA (86.8152 \%)                      & Age Band (93.0241 \%)                             \\
        Gender (89.6640 \%)                     & VLE Clicks: forumng (85.9021 \%)                  & Disability (86.8517 \%)                           \\
        VLE Clicks: homepage (89.2988 \%)       & VLE Clicks: oucontent (81.3733 \%)                & Assessment: TMA (86.7421 \%)                      \\
        Region (88.8605 \%)                     & Num Of Prev Attempts (81.3002 \%)                 & VLE Clicks: forumng (85.9386 \%)                  \\
        Module Presentation Length (86.1943 \%) & Imd Band (80.0950 \%)                             & Gender (80.3871 \%)                               \\
        Assessment: TMA (85.1717 \%)            & Disability (78.0862 \%)                           & VLE Clicks: subpage (76.9175 \%)                  \\
        Imd Band (82.9803 \%)                   & VLE Clicks: subpage (76.2966 \%)                  & VLE Clicks: oucontent (76.2966 \%)                \\
        Disability (82.7611 \%)                 & Studied Credits (72.5712 \%)                      & Code Presentation (71.5851 \%)                    \\
        Date Registration (74.5800 \%)          & Code Presentation (71.2564 \%)                    & Region (71.5486 \%)                               \\
        VLE Clicks: forumng (74.3608 \%)        & VLE Clicks: resource (70.8546 \%)                 & VLE Clicks: resource (70.0877 \%)                 \\
        Studied Credits (74.0687 \%)            & Date Registration (66.1066 \%)                    & Imd Band (65.3762 \%)                             \\
        VLE Clicks: oucontent (70.9277 \%)      & Age Band (62.6370 \%)                             & Highest Education (64.3170 \%)                    \\
        Highest Education (70.8912 \%)          & Region (59.7882 \%)                               & Num Of Prev Attempts (62.0891 \%)                 \\
        VLE Clicks: resource (69.7589 \%)       & Gender (57.4142 \%)                               & Assessment: CMA (52.7757 \%)                      \\
        VLE Clicks: subpage (65.1570 \%)        & Assessment: CMA (53.1045 \%)                      & Code Module (52.7757 \%)                          \\
        Num Of Prev Attempts (63.4405 \%)       & Highest Education (52.4105 \%)                    & Studied Credits (52.1914 \%)                      \\
        VLE Clicks: quiz (50.5844 \%)           & VLE Clicks: quiz (51.9357 \%)                     & VLE Clicks: quiz (49.6713 \%)                     \\
        Assessment: CMA (48.4660 \%)            & Module Presentation Length (49.5617 \%)           & Module Presentation Length (47.4069 \%)           \\
        VLE Clicks: url (46.4573 \%)            & Code Module (48.8678 \%)                          & Date Registration (46.5668 \%)                    \\
        Code Module (42.7684 \%)                & VLE Clicks: url (47.6991 \%)                      & VLE Clicks: url (42.5858 \%)                      \\
        VLE Clicks: oucollaborate (20.4164 \%)  & VLE Clicks: oucollaborate (23.5573 \%)            & VLE Clicks: oucollaborate (19.8320 \%)            \\
        VLE Clicks: page (18.1885 \%)           & VLE Clicks: ouwiki (21.1468 \%)                   & Assessment: Exam (17.3119 \%)                     \\
        Assessment: Exam (17.3119 \%)           & VLE Clicks: page (19.4668 \%)                     & VLE Clicks: externalquiz (16.6545 \%)             \\
        VLE Clicks: ouwiki (15.8145 \%)         & VLE Clicks: glossary (18.8459 \%)                 & VLE Clicks: page (16.4354 \%)                     \\
        VLE Clicks: externalquiz (14.7188 \%)   & Assessment: Exam (17.3119 \%)                     & VLE Clicks: ouwiki (15.8510 \%)                   \\
        VLE Clicks: dataplus (9.6056 \%)        & VLE Clicks: externalquiz (16.6910 \%)             & VLE Clicks: glossary (15.7779 \%)                 \\
        VLE Clicks: glossary (8.3638 \%)        & VLE Clicks: dualpane (10.4821 \%)                 & VLE Clicks: dualpane (10.5917 \%)                 \\
        VLE Clicks: dualpane (7.8890 \%)        & VLE Clicks: questionnaire (8.4368 \%)             & VLE Clicks: questionnaire (7.2316 \%)             \\
        VLE Clicks: questionnaire (6.8298 \%)   & VLE Clicks: folder (6.5741 \%)                    & VLE Clicks: folder (6.5741 \%)                    \\
        VLE Clicks: folder (6.5011 \%)          & VLE Clicks: ouelluminate (4.1636 \%)              & VLE Clicks: ouelluminate (2.5566 \%)              \\
        VLE Clicks: ouelluminate (3.6888 \%)    & VLE Clicks: dataplus (2.4836 \%)                  & VLE Clicks: htmlactivity (2.0088 \%)              \\
        VLE Clicks: htmlactivity (2.3375 \%)    & VLE Clicks: htmlactivity (2.4470 \%)              & VLE Clicks: dataplus (1.9722 \%)                  \\
    \end{tabular}
    \caption{Common instance types sorted by bag presence in percent for the two extreme preference settings 0 and 1 of the task-agnostic hypernetwork and the baseline method using mean pooling.}
    \label{tab:topk}
\end{table*}

Table \ref{tab:topk} shows the common instance types sorted by the fraction of bags they are present in for the baseline model and the two extreme configurations of the task-agnostic hypernetwork framework. Overall, relative frequencies of instances change only slightly and not in a systematic way. Especially the four features defined as protected do not differ strongly in presence between the two hypernetwork configurations.

The sparsity and consistency metrics for all models and all pooling methods show that the model explanations are comparably sparse and very consistent with both Spearman's $\rho$ and Jaccard's K close to 1 for all entries. They are not different for different preference values.

\begin{table*}
    \centering
    \begin{tabular}{l|l|l|r|r|r|r|r}
        \multicolumn{3}{c}{} & \multicolumn{2}{c}{Sparsity} & \multicolumn{2}{c}{Consistency} & \\
        \cmidrule(r){4-5} \cmidrule(l){6-7}
        \textbf{Model}             & \textbf{Preference} & \textbf{Pooling} & \textbf{Gini Coeff.} & \textbf{HHI Score} & \textbf{Spearman's $\rho$} & \textbf{Jaccard (K=5)} & \textbf{Demographics (\%)} \\
        \hline
        Task-agnostic & 0.0                 & Rep-The-Set      & 0.4793                    & 0.0614             & 0.9983                               & 1.0000                 & 12.0\%                                 \\
        Task-agnostic & 1.0                 & Rep-The-Set      & 0.4796                    & 0.0613             & 0.9972                               & 1.0000                 & 12.0\%                                 \\
        Task-aware    & 0.0                 & Rep-The-Set      & 0.4939                    & 0.0559             & 0.9980                               & 0.6667                 & 15.2\%                                 \\
        Task-aware    & 1.0                 & Rep-The-Set      & 0.4932                    & 0.0558             & 0.9985                               & 0.6667                 & 15.2\%                                 \\
        Task-agnostic & 0.0                 & Attention        & 0.5275                    & 0.0621             & 0.9982                               & 1.0000                 & 18.7\%                                 \\
        Task-agnostic & 1.0                 & Attention        & 0.5376                    & 0.0658             & 0.9938                               & 1.0000                 & 18.1\%                                 \\
        Task-aware    & 0.0                 & Attention        & 0.4862                    & 0.0539             & 0.9988                               & 1.0000                 & 15.6\%                                 \\
        Task-aware    & 1.0                 & Attention        & 0.4744                    & 0.0512             & 0.9975                               & 0.6667                 & 16.7\%                                 \\
        Task-agnostic & 0.0                 & Max              & 0.5134                    & 0.0602             & 0.9988                               & 1.0000                 & 13.9\%                                 \\
        Task-agnostic & 1.0                 & Max              & 0.5316                    & 0.0641             & 0.9970                               & 1.0000                 & 13.1\%                                 \\
        Task-aware    & 0.0                 & Max              & 0.5071                    & 0.0623             & 0.9985                               & 1.0000                 & 10.6\%                                 \\
        Task-aware    & 1.0                 & Max              & 0.5071                    & 0.0622             & 0.9983                               & 1.0000                 & 10.6\%                                 \\
        Baseline                   &                     & Mean             & 0.4470                    & 0.0507             & 0.9961                               & 0.6667                 & 18.0\%                                 \\
        Task-agnostic & 0.0                 & Mean             & 0.4531                    & 0.0591             & 0.9982                               & 1.0000                 & 13.0\%                                 \\
        Task-agnostic & 1.0                 & Mean             & 0.4886                    & 0.0611             & 0.9936                               & 1.0000                 & 15.5\%                                 \\
        Task-aware    & 0.0                 & Mean             & 0.4907                    & 0.0654             & 0.9976                               & 1.0000                 & 13.1\%                                 \\
        Task-aware    & 1.0                 & Mean             & 0.4940                    & 0.0655             & 0.9980                               & 1.0000                 & 13.6\%                                 \\
    \end{tabular}
    \caption{Sparsity, Consistency and Demographics Reliance metrics for several model configurations. Baseline values are only available for mean pooling due to unexpected maintenance on the Snellius national supercomputer.}
    \label{tab:metrics}
\end{table*}

\section{Discussion}
\label{sec:discussion}

The results show that the proposed hypernetwork frameworks do not provide a reliable mechanism for continuously controlling the severity of demographic bias. This section interprets that failure, examines the optimization and state--action-space mechanisms that may have produced mode collapse, and outlines alternative directions for fair and interpretable RL--MIL. Ethical implications of using student-at-risk predictions are considered at the end.

\subsection{Elaboration}

Because both proposed frameworks suffer from mode collapse, the experiments do not establish a meaningful controllable fairness--performance frontier. Instead, the evidence indicates strong objective dominance and/or vanishing gradients in the hypernetwork, with preference-conditioned configurations converging toward essentially the same behavior. Since the models fail to adapt to different preference configurations, it is impossible to say how instance selection policies and model explanations change in low-bias environments. The other important metrics of reproducibility and reliability are technically solid, but because of the missing validity regarding the goals of this study that only reflects the model's ability to perform similar to the baseline. All in all, the hypernetwork frameworks do not offer any value compared to the baseline model. In the following, we list some of the possible reasons for the mode collapse of both hypernetwork architectures and propose directions for future research.

\subsection{Hyperparameter grid}

It is possible that succeeding in this avenue requires more consideration regarding the hyperparameters. We did not adjust the learning rate for the different parts of the framework separately, which is very likely a mistake. Learning to recognize protected features in the final layer of the task model's MLP is a much simpler task than learning how to adjust a model's parameters to reduce bias. Our hyperparameter grid should therefore have included separate learning rate ranges for the task model, the debiasing model, the policy network and the hypernetwork.

\subsection{Complex implications of small parameter changes}

The hypernetwork choosing parameter configurations for the different models based on a preference for performance or low bias needs to learn complex nonlinear patterns between the parameters. Also, small changes of the parameters lead to big implications for the selection of instances, which makes this task even more difficult. We tried to account for this using Fourier Embeddings for our preference scalar, but that was possibly not enough. To tackle this issue, one could think of adding another hidden layer or limiting the output parameter space softly using transformations.

\subsection{State-action space shift}

Training all parts of the model at the same time could have led to an uncertain state-action space. A task model adapting to the outputs of a debiasing model (task-aware hypernetwork framework) changes the state space more severely than if it had just been trained on the performance signal. This added uncertainty is propagated to the hypernetwork, since the state-action space is different for each preference value. It is possible that this level of uncertainty eliminates any potential of learning the impacts of parameter adjustments. Mitigating this issue entails carefully adjusting hyperparameters, network structures and the training regimen itself, maybe even excluding parts of the model from adaptive training alongside the hypernetwork.

\subsection{Future work}

Future work could address the optimization failures directly through alternative conditioning mechanisms, gradient-balancing strategies, or separate objective-specific parameterizations. A second direction is to reformulate the debiasing problem through causal regularization rather than relying exclusively on adversarial pressure.

\tikzstyle{var} = [rectangle, rounded corners, minimum width=1.5cm, minimum height=0.75cm,text centered, draw=black]
\tikzstyle{arrow} = [thick,->,>=stealth]
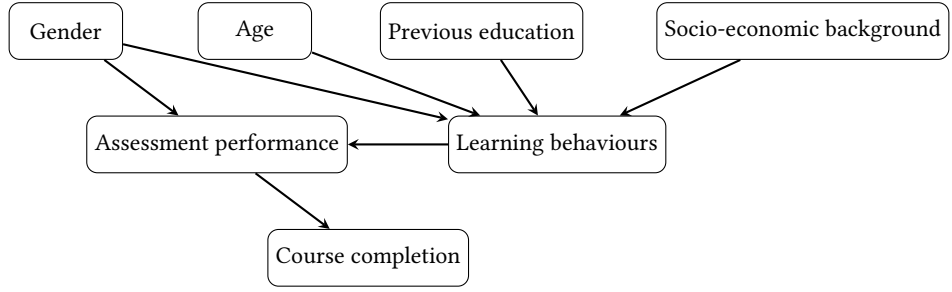
\begin{figure*}[ht]
\begin{tikzpicture}[node distance=1.5cm]
\node (g) [var] {Gender};
\node (a) [var, right of=g, xshift=1cm] {Age};
\node (e) [var, right of=a, xshift=1.5cm] {Previous education};
\node (s) [var, right of=e, xshift=2.75cm] {Socio-economic background};
\node (t) [var, below of=g, xshift=2cm] {Assessment performance};
\node (l) [var, right of=t, xshift=3cm] {Learning behaviours};
\node (o) [var, below of=t, xshift=2cm] {Course completion};
\draw [arrow] (g) -- (t);
\draw [arrow] (g) -- (l);
\draw [arrow] (a) -- (l);
\draw [arrow] (e) -- (l);
\draw [arrow] (s) -- (l);
\draw [arrow] (l) -- (t);
\draw [arrow] (t) -- (o);
\end{tikzpicture}
\Description{A schematic display of the structural causal model we will use to adjust classification reward strength.}
\caption{Schematic depiction of the expected causal patterns between observed variables in our problem setting. Arrows point from causal ancestors to their children.}
\label{fig:scm}
\end{figure*}

To examine and guard against more complex issues introducing bias into a classification model's predictions like spurious correlations, the extensive literature on causality provides helpful insight. \cite{deng_causal_2023} summarize different approaches that connect causality research and RL methods for several purposes. Using the idea of a Structural Causal Model (SCM) to describe the relationships between (un)observed variables in a given domain, interventions can be administered to find out whether the proposed model is accurate to the generated data. This insight can then be used for example to regularize an RL agent \cite{deng_causal_2023}.

This direction is consistent with recent evidence that adversarial removal of demographic information can reduce sensitive-attribute leakage but may also degrade task performance, whereas causal bottlenecks can strengthen demographic separation at an even larger utility cost \cite{olijslager_causally_2026}. Likewise, RLMIL-DAT demonstrates that adversarial objectives can be coupled directly to reinforced instance selection when the nuisance signal is domain identity \cite{shu_crosslanguage_2026}; the present results indicate that controlling a fairness objective through a preference-conditioned hypernetwork is substantially less stable.

One example of such an intervention is frontdoor/backdoor adjustment. Based on the application of the do-operator to remove a variable from its causal parents, spurious correlations between a treatment variable, its confounders and the output variable are blocked after frontdoor/backdoor adjustment \cite{scholkopf_toward_2021}.

The classification signal involved in the training of our framework can suffer from spurious correlation. To mitigate the observed relationships, future researchers could administer backdoor adjustment as formalized in the causal-inference literature \cite{bareinboim_pearls_2022,geffner_probabilistic_2022}. We propose a Structural Causal Model (SCM) based on educational psychological research to this end. Whether a student passes or fails a course is decided by their assessment performance. Thus, we suspect that assessment performance is a causal ancestor of course completion. How well students perform on assessments in turn is influenced by their studying behaviors. Besides studied facts, strategic adaptation to an exam's format can improve performance as well, for example by employing a process of elimination on multiple choice questions. As test taking strategies can change the result of an examination greatly and there appears to exist a gender bias for the kind and intensity of using these strategies, we claim that gender serves as a causal ancestor to assessment performance as well \cite{vermunt_patterns_2004}. Previous education and socio-economic background (represented by the hometown in the OULA dataset) have been shown to not influence assessment performance in addition to the effect of a student's grade point average \cite{amankwaa_previous_2015}. A direct link between a person's age and test-taking strategies or assessment performance in general has not been established yet, and there is a lack of evidence for indirect effects as older students should, for example, have more test-taking experience. This is why we do not include age as a causal ancestor to assessment performance. All mentioned protected variables, however, could influence the studying behavior of students as summarized by \cite{stenlund_group_2017}. A schematic depiction of the proposed SCM can be seen in Figure \ref{fig:scm}. Performing backdoor adjustment to disentangle the reward $R$ for the classification of students being at-risk $Y$ from the relationship between the suspected mediator variables $A$ (learning behaviours and assessment performance) and their protected ancestor variables $P$ (age, gender, educational background, socio-economic status) can be carried out according to:
\begin{align}
&\frac{E(Y=y|\text{do}(A=a))}{E(Y=y|A=a)} \\
&= \frac{\sum_p E(Y=y|A=a, P=p) \cdot P(P=p)}{\sum_p E(Y=y|A=a, P=p) \cdot P(A=a|P=p) \cdot P(P=p)}\\
&= \sum_p \frac{1}{P(A=a|P=p)} \\
R_{adj}&=\frac{R}{P(A=a|P=p)} \\
\end{align}.

\subsection{Ethical considerations}

When using the advancements in the research field of student-at-risk prediction for designing interventions in a real-life setting, important ethical considerations have to be made. Most importantly, such a system should only include machine learning techniques as suggestion generators with humans deciding which intervention should be pursued. It is also crucial that the used method does not suffer from severe bias as the explored techniques currently do. At the same time, too many false positives or false negatives have to be avoided as well to prevent resource misallocation and distrust for this socially delicate matter. It is unclear whether an acceptable algorithm meeting all these requirements can be built. Resolving the highly biased nature of many methods while still securing interpretability of decisions should be the main focus of future research.

\section{Conclusion}
\label{sec:conclusion}
This study extends RL--MIL student-at-risk prediction with preference-conditioned hypernetworks intended to control the trade-off between predictive performance and demographic bias. The evaluation tests whether adversarial debiasing can produce a configurable fairness--performance frontier and whether increasing fairness pressure changes the learned instance-selection policy.

Our results show that while we were able to replicate the findings of the previous work, our proposed novel network structure failed to capture the patterns we wanted to examine. This finding could have many different reasons, as this endeavor entails many complexities. Carefully adjusting hyperparameter spaces, designing different hypernetwork architectures and/or changing the training regimen could alleviate these issues.

Overall, the proposed hypernetwork design is not suitable for reliable preference-controlled fairness optimization in its current form. The negative result is nevertheless informative: adding a conditioning scalar is insufficient when gradients from the competing objectives do not produce distinct parameter regimes. Future work should therefore treat optimization stability and objective separation as first-class design requirements. However, it might also be fruitful to take a different approach to training a hypernetwork. For example, future work could focus on administering causal regularization to see whether this simple method is already enough to reduce model bias.

\bibliographystyle{ACM-Reference-Format}
\bibliography{bibtex_acm_updated}

\onecolumn

\appendix
\small
\setlength{\tabcolsep}{4pt}

\section{Plots and use of generative AI}
\label{sec:apx:A}

Generative AI was used for this project in order to reason about PyTorch coding decisions, check grammar and spelling mistakes of this document and discover missing references.

\begin{figure*}[ht]
\includegraphics[width=\textwidth]{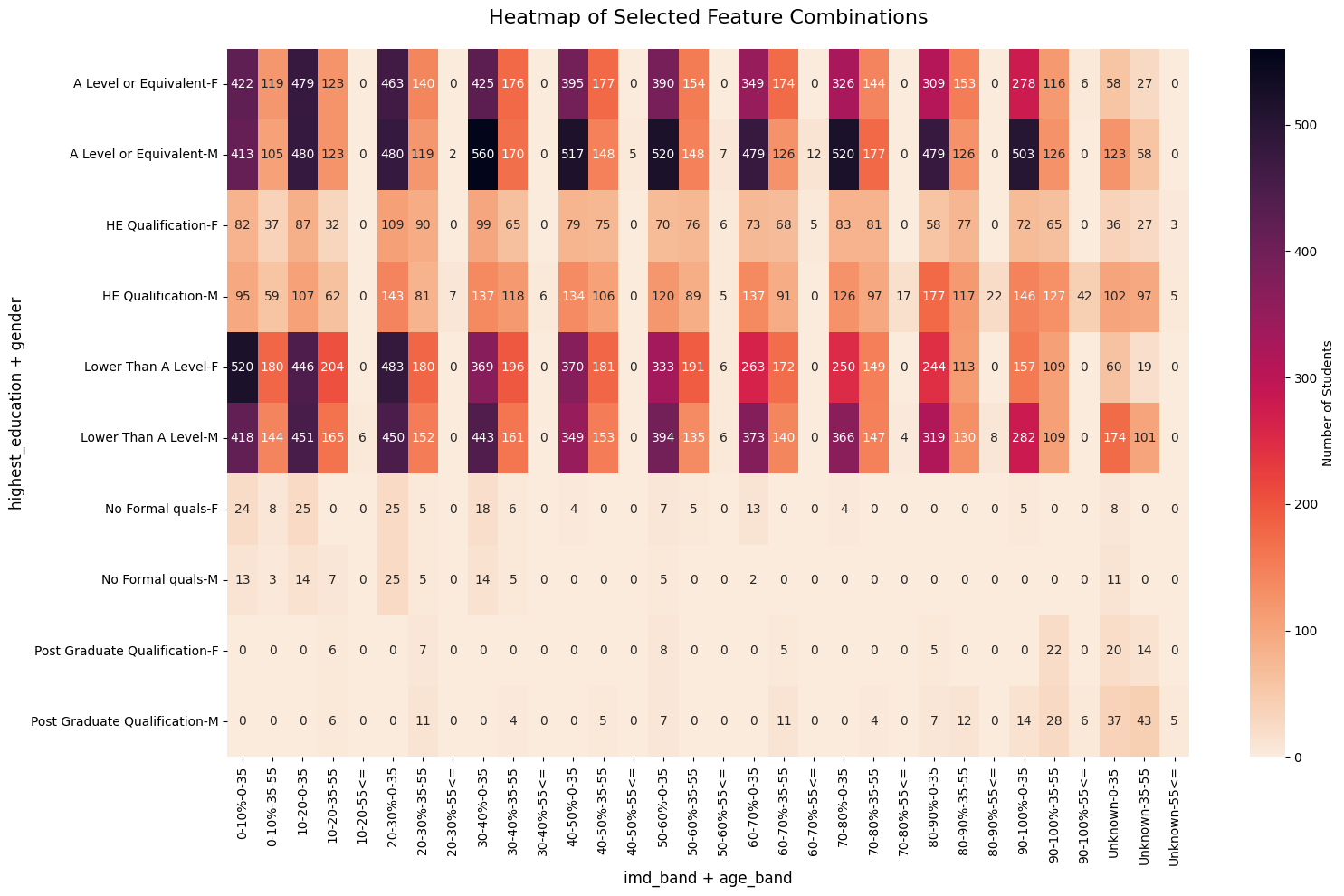}
\Description{TODO}
\caption{Number of students for each combination of protected features.}
\label{fig:class_heat}
\end{figure*}

\begin{figure*}[ht]
\includegraphics[width=\textwidth]{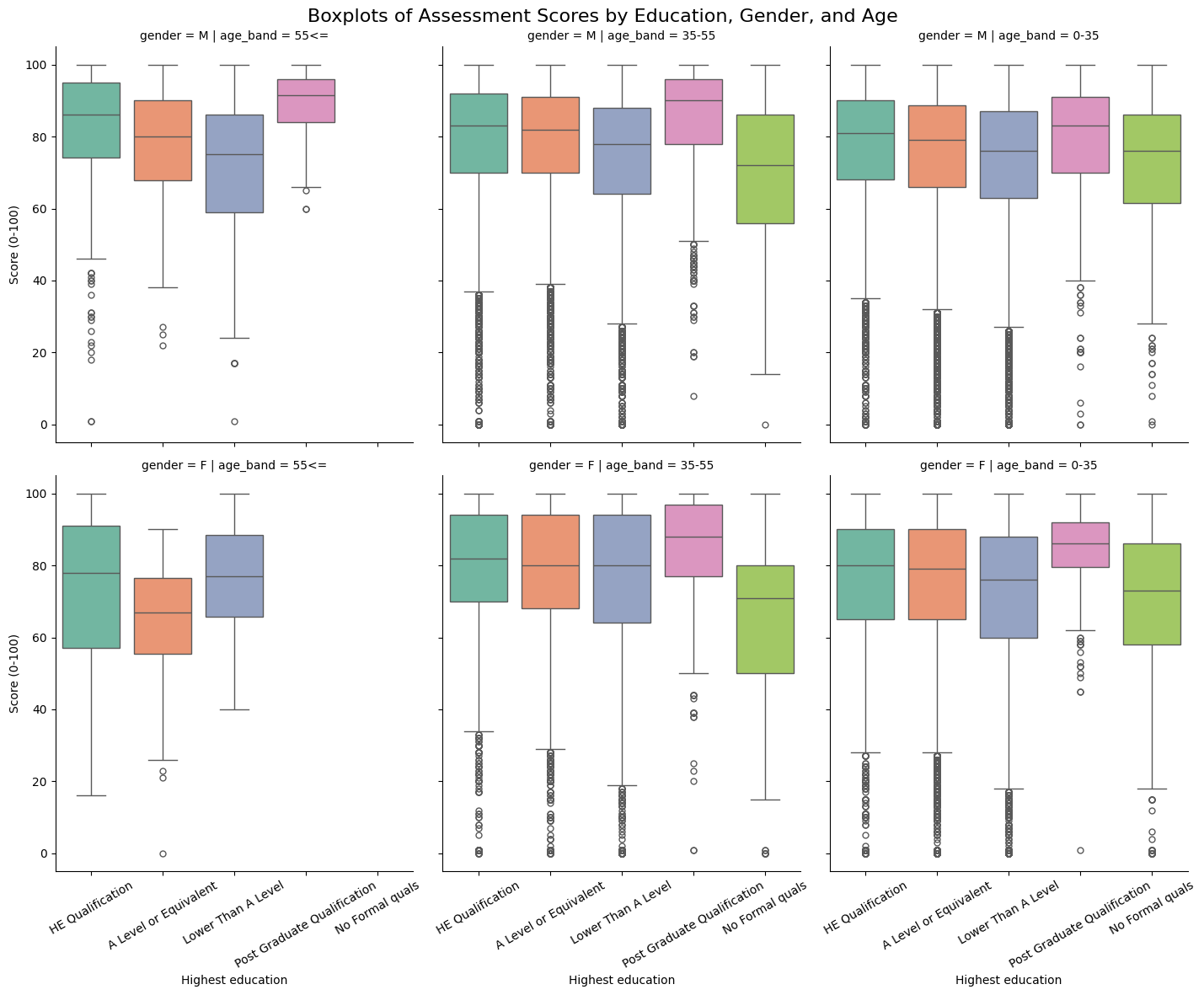}
\Description{TODO}
\caption{Summary plots of assessment scores separated by the protected features education, gender and age. The most noticeable relationship can be seen between education and assessment score, with post graduate qualification performing better and no formal qualifications performing worse than all other levels. For age and gender, no clear pattern emerges.}
\label{fig:assessment}
\end{figure*}

\begin{figure*}[ht]
\includegraphics[width=\textwidth]{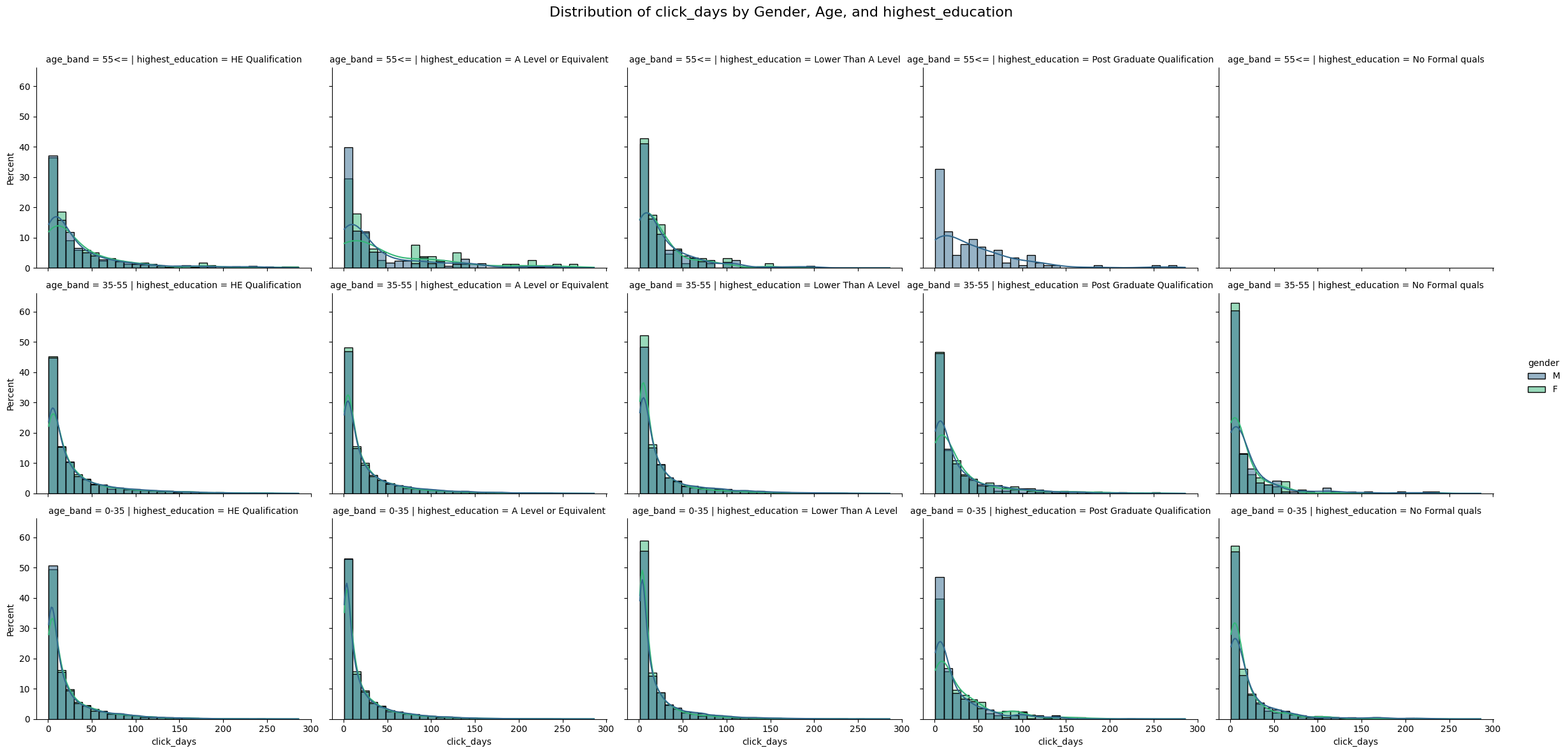}
\Description{TODO}
\caption{Distribution of the number of click days for each resource, split by gender, age and education. Again, the most clear but very subtle pattern can be seen by comparing the peak of the no formal qualifications level against the peak of all other education levels. For the mentioned level, many resources are clicked on less often than for other levels.}
\label{fig:vle}
\end{figure*}

\begin{figure*}[ht]
\includegraphics[width=\textwidth]{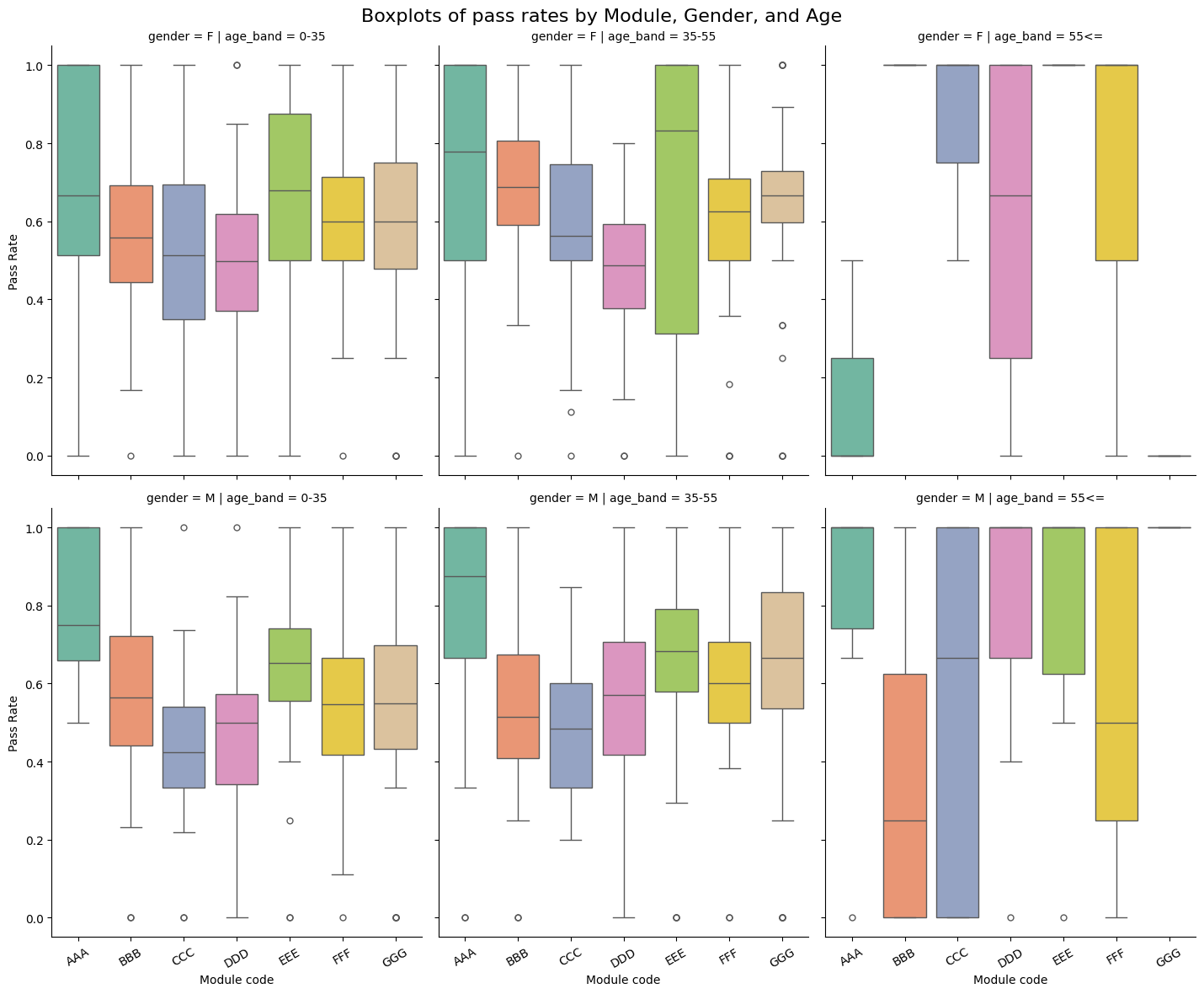}
\Description{TODO}
\caption{Summary plots of pass rates split by module, gender and age. Although there are stark differences between the courses, there is no clear effect for either age or gender.}
\label{fig:pass_rates}
\end{figure*}

\clearpage

\section{Tables}
\label{sec:apx:B}

\begin{longtable}{l|l|r|r|r}
    \centering
        \textbf{Framework}         & \textbf{Pooling} & \textbf{Preference} & \textbf{F1 score (performance) ($\pm$ SD)} & \textbf{Equalized Odds (bias) ($\pm$ SD)} \\
        \hline
        Baseline                   & Attention        &                     & 0.9245 ($\pm$ 0.0018)                      & 0.1681 ($\pm$ 0.0044)                     \\
        Baseline                   & Max              &                     & 0.9187 ($\pm$ 0.0012)                      & 0.0998 ($\pm$ 0.0288)                     \\
        Baseline                   & Mean             &                     & 0.9228 ($\pm$ 0.0017)                      & 0.1055 ($\pm$ 0.0160)                     \\
        Baseline                   & Rep-The-Set      &                     & 0.9280 ($\pm$ 0.0010)                      & 0.1426 ($\pm$ 0.0237)                     \\
        Task-agnostic Hypernetwork & Attention        & 0.0                 & 0.9261 ($\pm$ 0.0011)                      & 0.1344 ($\pm$ 0.0007)                     \\
        Task-agnostic Hypernetwork & Attention        & 0.1                 & 0.9266 ($\pm$ 0.0018)                      & 0.1408 ($\pm$ 0.0053)                     \\
        Task-agnostic Hypernetwork & Attention        & 0.2                 & 0.9252 ($\pm$ 0.0015)                      & 0.1573 ($\pm$ 0.0070)                     \\
\end{longtable}

\begin{longtable}{l|l|r|r|r|r}
    \centering
        \textbf{Framework}         & \textbf{Pooling} & \textbf{Seed} & \textbf{Preference} & \textbf{F1 score (performance) ($\pm$ SD)} & \textbf{Equalized Odds (bias) ($\pm$ SD)} \\
        \hline
        Baseline                   & Mean             & 0             &                     & 0.9230 ($\pm$ 0.0005)                      & 0.1019 ($\pm$ 0.0656)                     \\
        Task-agnostic Hypernetwork & Mean             & 0             & 0.0                 & 0.9323 ($\pm$ 0.0015)                      & 0.1690 ($\pm$ 0.0010)                     \\
        Task-agnostic Hypernetwork & Mean             & 0             & 0.1                 & 0.9301 ($\pm$ 0.0015)                      & 0.1711 ($\pm$ 0.0011)                     \\
        Task-agnostic Hypernetwork & Mean             & 0             & 0.2                 & 0.9301 ($\pm$ 0.0011)                      & 0.1691 ($\pm$ 0.0003)                     \\
        Task-agnostic Hypernetwork & Mean             & 0             & 0.3                 & 0.9316 ($\pm$ 0.0030)                      & 0.1721 ($\pm$ 0.0020)                     \\
        Task-agnostic Hypernetwork & Mean             & 0             & 0.4                 & 0.9319 ($\pm$ 0.0015)                      & 0.1707 ($\pm$ 0.0018)                     \\
        Task-agnostic Hypernetwork & Mean             & 0             & 0.5                 & 0.9308 ($\pm$ 0.0019)                      & 0.1722 ($\pm$ 0.0004)                     \\
\end{longtable}

\begin{table*}[ht]
    \centering
    \begin{tabular}{l|l|r|l}
        \textbf{Pooling} & \textbf{Tested Hypothesis}                           & \textbf{T value} & \textbf{p value} \\
        \hline
        Mean             & Task-agnostic Hypernetwork > Baseline                & 1.6058           & 0.1143           \\
        Mean             & Task-aware Hypernetwork > Baseline                   & 2.5671           & 0.0234 (*)       \\
        Mean             & Task-aware Hypernetwork > Task-agnostic Hypernetwork & 1.1183           & 0.2681           \\
        Max              & Task-agnostic Hypernetwork > Baseline                & 1.3277           & 0.1900           \\
        Max              & Task-aware Hypernetwork > Baseline                   & 2.9425           & 0.0114 (*)       \\
        Max              & Task-aware Hypernetwork > Task-agnostic Hypernetwork & 1.6701           & 0.1003           \\
        Attention        & Baseline > Task-agnostic Hypernetwork                & 0.8206           & 0.4156           \\
        Attention        & Task-aware Hypernetwork > Baseline                   & 0.2212           & 0.8284           \\
        Attention        & Task-aware Hypernetwork > Task-agnostic Hypernetwork & 1.2770           & 0.2067           \\
        Rep-The-Set      & Task-agnostic Hypernetwork > Baseline                & 1.4597           & 0.1503           \\
        Rep-The-Set      & Task-aware Hypernetwork > Baseline                   & 1.8108           & 0.0758           \\
        Rep-The-Set      & Task-aware Hypernetwork > Task-agnostic Hypernetwork & 0.3888           & 0.6983           \\
    \end{tabular}
    \caption{T test results of Equalized Odds Deviation for every combination of frameworks and each pooling method.}
    \label{tab:bias_ttests}
\end{table*}

\begin{table*}[ht]
    \centering
    \begin{tabular}{l|l|r|l}
        \textbf{Pooling} & \textbf{Tested Hypothesis}                           & \textbf{T value} & \textbf{p value} \\
        \hline
        Mean             & Baseline > Task-agnostic Hypernetwork                & 0.2529           & 0.8013           \\
        Mean             & Task-aware Hypernetwork > Baseline                   & 3.3349           & 0.0054 (*)       \\
        Mean             & Task-aware Hypernetwork > Task-agnostic Hypernetwork & 1.4354           & 0.1566           \\
        Max              & Task-agnostic Hypernetwork > Baseline                & 0.6903           & 0.4930           \\
        Max              & Baseline > Task-aware Hypernetwork                   & 8.1000           & 0.0000 (*)       \\
        Max              & Task-agnostic Hypernetwork > Task-aware Hypernetwork & 7.2982           & 0.0000 (*)       \\
        Attention        & Task-agnostic Hypernetwork > Baseline                & 0.0741           & 0.9412           \\
        Attention        & Task-aware Hypernetwork > Baseline                   & 7.2516           & 0.0000 (*)       \\
        Attention        & Task-aware Hypernetwork > Task-agnostic Hypernetwork & 12.1674          & 0.0000 (*)       \\
        Rep-The-Set      & Baseline > Task-agnostic Hypernetwork                & 0.9965           & 0.3235           \\
        Rep-The-Set      & Baseline > Task-aware Hypernetwork                   & 1.1885           & 0.2399           \\
        Rep-The-Set      & Task-agnostic Hypernetwork > Task-aware Hypernetwork & 0.5061           & 0.6139           \\
    \end{tabular}
    \caption{T test results of F1-score (classification performance) for every combination of frameworks and each pooling method.}
    \label{tab:perf_ttests}
\end{table*}

\begin{table*}[ht]
    \centering
    \begin{tabular}{l|l|l|r|l}
        \textbf{Pooling} & \textbf{Framework}         & \textbf{Tested Hypothesis}  & \textbf{T value} & \textbf{p value} \\
        \hline
        Mean             & Task-agnostic Hypernetwork & Preference 0 > Preference 1 & -0.9422          & 0.3737           \\
        Mean             & Task-aware Hypernetwork    & Preference 0 > Preference 1 & -0.8843          & 0.4698           \\
        Max              & Task-agnostic Hypernetwork & Preference 0 > Preference 1 & 0.0100           & 0.9922           \\
        Max              & Task-aware Hypernetwork    & Preference 0 > Preference 1 & -0.2726          & 0.8107           \\
        Attention        & Task-agnostic Hypernetwork & Preference 0 > Preference 1 & -0.0245          & 0.9811           \\
        Attention        & Task-aware Hypernetwork    & Preference 0 > Preference 1 & -0.9305          & 0.4503           \\
        Rep-The-Set      & Task-agnostic Hypernetwork & Preference 0 > Preference 1 & -0.6391          & 0.5406           \\
        Rep-The-Set      & Task-aware Hypernetwork    & Preference 0 > Preference 1 & 0.8029           & 0.4452           \\
    \end{tabular}
    \caption{T test of Equalized Odds Deviation between preference 0 and preference 1 for both hypernetwork architectures and all pooling methods.}
    \label{tab:pareto_bias}
\end{table*}

\end{document}